\pdfoutput=1
\documentclass[11pt]{article}

\usepackage[final]{emnlp2022}
\usepackage{times}
\usepackage{latexsym}

\usepackage[ruled]{algorithm2e}
\usepackage{booktabs}
\usepackage{placeins}
\usepackage{afterpage}
\usepackage{times}
\usepackage{latexsym}
\usepackage{graphicx}
\usepackage{multirow}
\usepackage{multicol}
\usepackage{hyperref}
\usepackage{cite}
\usepackage{makecell}
\usepackage[T1]{fontenc}
\usepackage[utf8]{inputenc}
\usepackage{microtype}
\usepackage[flushleft]{threeparttable}
\usepackage{amsmath,amssymb,amsfonts}
\usepackage{scalerel,xparse}

\usepackage{microtype}

\DeclareMathOperator*{\MHAtt}{MHAtt}
\DeclareMathOperator*{\Att}{Att}

\title{Shapley Head Pruning: \\Identifying and Removing Interference in Multilingual Transformers}

\author{William Held \\
  \texttt{wheld3@gatech.edu} \\\And
  Diyi Yang \\
  \texttt{diyiy@stanford.edu} \\ }

\date{}

\begin{document}
\maketitle

\abovedisplayskip 7pt plus2pt minus5pt%
\belowdisplayskip \abovedisplayskip
\abovedisplayshortskip  0pt plus3pt%
\belowdisplayshortskip  4pt plus3pt minus3pt%

\begin{abstract}
  Multilingual transformer-based models demonstrate remarkable zero and few-shot transfer across languages by learning and reusing language-agnostic features. However, as a fixed-size model acquires more languages, its performance across all languages degrades, a phenomenon termed interference. Often attributed to limited model capacity, interference is commonly addressed by adding additional parameters despite evidence that transformer-based models are overparameterized. In this work, we show that it is possible to reduce interference by instead identifying and pruning language-specific parameters. First, we use Shapley Values, a credit allocation metric from coalitional game theory, to identify attention heads that introduce interference. Then, we show that removing identified attention heads from a fixed model improves performance for a target language on both sentence classification and structural prediction, seeing gains as large as 24.7\%. Finally, we provide insights on language-agnostic and language-specific attention heads using attention visualization.
\end{abstract}
  
\section{Introduction}
\begin{figure*}
\begin{center}
\includegraphics[width=\paperwidth-5cm]{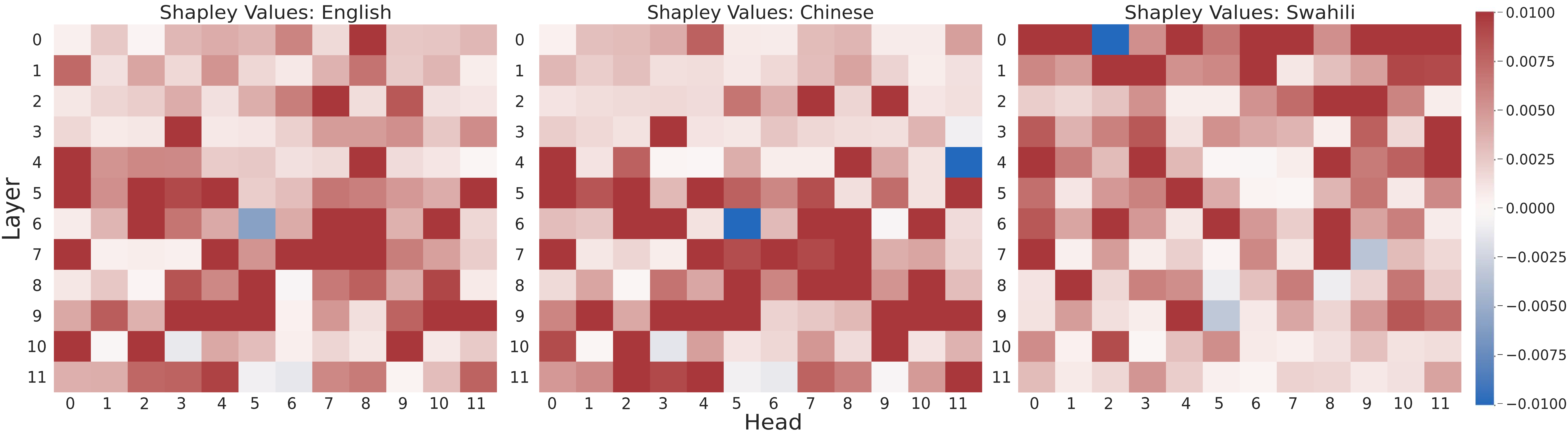}
\caption{Attention Head Shapley Values of 3 Languages for XLM-R finetuned on English XNLI. Each value indicates the mean marginal effect an attention head has on accuracy for the XNLI test set in the language.}\label{figure2}
\end{center}
\end{figure*}

Cross-lingual transfer learning aims to utilize a natural language processing system trained on a source language to improve results for the same task in a different target language. The core goal is to maintain relevant learned patterns from the source while disregarding those which are inapplicable to the target. Multilingual pretraining of transformer language models has recently become a widespread method for cross-lingual transfer; demonstrating remarkable zero and few shot performance across languages when finetuned on monolingual data~\citep{mbert, xlmr, mt5-xue}.  

However, adding languages beyond a threshold begins to harm cross-lingual transfer in a fixed-size model as shown in prior work~\citep{xlmr, mt5-xue}. This has been addressed with additional parameters, both language-specific~\citep{pfeiffer-etal-2020-mad} and broadly~\citep{xlmr,mt5-xue}. \citet{wang-etal-2020-negative} justifies this by showing that competition over limited capacity drives interference. However, limited capacity seems unlikely to be the sole driver as \citet{lotterytransformers} shows that pretrained language models are highly overparametrized. 

We offer an alternate hypothesis that interference is caused by components that are specialized to language-specific patterns and introduce noise when applied to other languages. To test this hypothesis, we introduce a methodology for identifying language-specific components and further removing them, which is expected to improve model performance without updating or adding additional language-specific parameters. Our work builds on prior research studying monolingual models which shows that they can be pruned aggressively~\citep{michel-16, voita-etal-2019-analyzing}. 

We leverage Shapley Values, the mean marginal contribution of a player to a collaborative reward, to identify attention heads that cause interference. Shapley Values are consistent, gradient-free which allows for binary removal, and map harmful components clearly to negative values. While the process of identifying and pruning language-specific structures is agnostic to attribution methodology, this makes Shapley Values particularly well-suited to the task. To make computation tractable, we follow prior work in vision models~\citep{neuronshap} to approximate Shapley Values more efficiently using truncation and multi-armed bandit sampling. We contribute the following:

\begin{enumerate}
    \item \textbf{Attention Head Language Affinity:} Even when computed from aligned sentences, Attention Head Shapley Values vary based on the language of input. This highlights that a subset of attention heads has language-specific importance, while others are language-agnostic as shown in Figure \ref{figure2}.
    
    \item \textbf{Improving through Pruning:} Model pruning according to Shapley Values improves performance without updating parameters on the Cross-Lingual Natural Language Inference corpus~\citep{conneau-etal-2018-xnli} and the Universal Dependencies Part-of-Speech corpus~\citep{udpos}. This opens a path of work to reduce interference by removing parameters rather than adding them.
    
    \item \textbf{Interpreting Multilingual Heads:} In a qualitative study, we find that the most language-agnostic heads identified have human interpretable language-agnostic function, while language-specific heads have varied behavior.
    
\end{enumerate}

\section{Related Work}
\subsection{Multilingual Learning}
A large amount of work has studied both the theoretical underpinnings of learning common structures for language and their applications to cross-lingual transfer. Early works exploited commonality through the use of pivot representations, created either by translation~\citep{ mann-yarowsky-2001-multipath, tiedemann-etal-2014-treebank, mayhew-etal-2017-cheap} or language-agnostic task formulations~\citep{zeman2008reusable, mcdonald2011multi}. 

As NLP has increasingly used representation learning, dense embedding spaces replaced explicit pivots. This led to methods that identified the commonalities of embedding spaces and ways to align them~\citep{joulin-etal-2018-loss, artetxe-etal-2018-robust, artetxe-schwenk-2019-massively}. Recently, many works have trained multilingual transformer models~\citep{mbert, xlmr, mbart, mt5-xue, hu-etal-2021-explicit} as the basis for cross-lingual transfer. These models both implicitly and explicitly perform alignment, although they empirically achieve stronger alignment between closely related languages~\citep{artetxe-etal-2020-cross, conneau-etal-2020-emerging}. 

With language-specific data, further work has studied how to reduce interference by adding a small number of language-specific parameters. These works adapt a model for the target language by training only Adapters~\citep{wang-etal-2020-negative, pfeiffer-etal-2020-mad, ansell-etal-2021-mad-g}, prompts~\citep{zhao-schutze-2021-discrete}, or subsets of model parameters~\citep{ansell-etal-2022-composable}.

\citet{ma-etal-2021-contributions} previously investigated pruning in multilingual models using gradient-based importance metrics to study variability across attention heads. However, they used a process of iterative pruning and language-specific finetuning. This iterative process is largely unstable since there are many trainable subnetworks within large models~\citep{prasanna-etal-2020-bert}. Our method is the first to address interference and improve cross-lingual performance purely by pruning, without updating or adding additional language-specific parameters.

\subsection{Model Pruning}
Model pruning has largely been focused reducing the onerous memory and computation requirements of large models. These techniques are broken into two approaches: structured and unstructured pruning. \emph{Unstructured pruning} aims to remove individual parameters, which allows for more fine-grained removal. This process often has minimal effects even at extremely high degrees of sparsity. To efficiently prune a large number of parameters, many techniques propose using gradients or parameter magnitude~\citep{integratedGradients, lee2018snip, lotteryticket, lotterytransformers} as importance metrics.

\emph{Structured pruning}, or removing entire structural components, is motivated by computational benefits from hardware optimizations. In the case of Transformers, most of this pruning work targets removal of attention heads, either through static ranking~\citep{michel-16} or through iterative training~\citep{voita-etal-2019-analyzing, prasanna-etal-2020-bert, compact}. These pruning methods have also been used to study model behavior, but methods with iterative finetuning are highly unstable as many sub-networks can deliver the same level of performance once trained~\citep{prasanna-etal-2020-bert}.

Our work studies pruning without updating model parameters, which aligns with ~\citet{michel-16} which was able to maintain reasonable model performance even when removing 50\% of total attention heads. Furthermore, \citet{kovaleva-etal-2019-revealing} found that pruning attention heads could sometimes improve model performance without further finetuning. We build on this to develop a methodology for consistently identifying pruned models which improve performance.

\section{Methods}
To identify and remove interference, we need a metric which can distinguish harmful, unimportant, and beneficial attention heads. Prior work~\citep{michel-16, ma-etal-2021-contributions} utilized the magnitude of gradients as an importance metric. However, this metric measures the sensitivity of the loss function to the masking of a particular head regardless of the direction of that sensitivity. Since the loss function is sensitive to the removal of both harmful and beneficial heads, we develop a simple yet effective method which separates these classes.

Shapley Values~\citep{shapley_1953} have often been applied in model interpretability since they are the only attribution method to abide by the theoretical properties of local accuracy, missingness, and consistency laid out by ~\citet{shap}. In our setting, Shapley Values have two advantages over gradient-based importance metrics. Firstly, gradient-based approaches require  differentiable relaxations of evaluation functions and masking, but Shapley Values do not. Therefore, we can instead use the evaluation functions and binary masks directly. Secondly, Shapley Values are meaningfully signed which allows them to distinguish beneficial, unimportant, and harmful heads rather than just important and unimportant heads. This latter property is essential for our goal of identifying interference.

We apply Shapley Values to the task of structural pruning. In order to compute Shapley Values for each head, we first formalize the forward pass of a Transformer as a coalitional game between attention heads. Then, we describe a methodology to efficiently approximate Shapley Values using Monte Carlo simulation combined with truncation and multi-armed bandit search. Finally, we propose a simple pruning algorithm using the resulting values to evaluate the practical utility of this theoretically grounded importance metric.

\subsection{Attention Head Shapley Values}
We formalize a Transformer performing a task as a coalitional game. Our set of players $A$ are attention heads of the model. In order to remove self-attention heads from the game without retraining, we follow \citet{michel-16} which augments multi-headed attention with an added gate $G_h=\{0, 1\}$ for each head $\Att_h$ in a layer with $N_h$ heads as follows:
    \begin{equation}
        \MHAtt(x, q) = \sum_{h=0}^{N_h} G_h \Att_h(x, q)
    \end{equation}
With $G_h=0$, that attention head does not contribute to the output of the transformer and is therefore considered removed from the active coalition.    
    
Our characteristic function $V(A)$ is the task evaluation metric $M_v(A)$ over a set of validation data within a target language, adjusted by the evaluation metric with all heads removed to abide by the $V(\emptyset) = 0$ property of coalitional games:
\begin{equation}
    V(A) = M_v(A) - M_v(\emptyset)
\end{equation}

With these established, the Shapley Value $\varphi_h$ for an attention head $Att_h$ is the mean performance improvement from switching gate $G_h$ from $0$ to $1$ across all $P$ permutations of other gates:

\begin{equation}
\varphi_h = \frac{1}{|P|}\sum_{A\in P} V(A\cup h) - V(A) \label{shapleyFormula}    
\end{equation}

\subsection{Approximating Shapley Values}\label{approx}
However, the exact computation of Shapley Values for $N$ attention heads requires $2^N$ evaluations of our validation metric which is intractable for the number of heads used in most architectures. This intractability is often addressed by using Monte Carlo simulation as an approximate~\citep{monte}. This replaces $P$ in Equation \ref{shapleyFormula} with randomly constructed permutations, rather than the full set of all possible solutions.

Computing low-variance Shapley Value estimates with Monte Carlo simulation alone still requires unreasonable amounts of compute time. Therefore, we follow \citet{neuronshap} to accelerate our computations. We add a truncation heuristic using priors about the behavior of neural networks and formulate estimation as a multi-armed bandit problem of separating harmful heads from all others. This reduces the number of samples required to compute consistent Shapley Values, with our experiments showing consistency even across languages.

\paragraph{Truncation Heuristics}
\noindent Truncation stops sampling the marginal contributions from the rest of a permutation of features once a stopping criterion is reached for that permutation of the Monte Carlo simulation. Prior work selects stopping criterion based on either total performance~\citep{neuronshap} or marginal improvements~\citep{datashap}. To avoid tailoring a threshold to each dataset, we instead choose to truncate based on the percentage of remaining attention heads. For all experiments, we truncate when less than 50\% of attention heads remain in the coalition to bias our estimations towards the effect of heads when the majority of the full network is present.

\paragraph{Multi-Armed Bandit Sampling}
\noindent The multi-armed bandit optimization stops sampling the marginal contributions of a particular player once a stopping criterion has been reached according to the variance of that player. This optimization uses Empirical Bernstein Bounds~\citep{empiricalbern} which establish limits on the true mean of a value using its variance. For $t$ samples with observed variance $\sigma_t$ and range $R$, there is a probability of $1-\delta$ that the difference between the observed mean $\hat{\mu}$ and true mean $\mu$ abides by the following inequality formulated by \citet{bernStopping}:

\begin{equation}
    |\hat{\mu} - \mu| \leq \sigma_t \sqrt{\frac{2\log(3/\delta)}{t}} + \frac{3R\log(3/\delta)}{t}
\label{bernEq}
\end{equation}

We stop sampling for a particular head once the lower bound established by this inequality is positive since this means there is a probability $1-\delta$ that this head is not harmful. This saves us significant computation without being likely to cause harmful attention heads to be missed. For all experiments in this paper, we use $R=1$ since the model's worst-case performance is zero and $\delta=0.1$ to give a 95\% confidence lower and upper bound.

\begin{table*}
\begin{center}
\begin{tabular}{|llllllllllllllll|}
\hline
Dataset                     & EN  & AR  & BG  & DE   & EL  & ES  & FR  & HI  & RU   & SW  & TH  & TR  & UR  & VI  & ZH  \\ \hline
\multicolumn{1}{|l|}{XNLI}  & 5.0 & 5.0 & 5.0 & 5.0  & 5.0 & 5.0 & 5.0 & 5.0 & 5.0  & 5.0 & 5.0 & 5.0 & 5.0 & 5.0 & 5.0 \\
\multicolumn{1}{|l|}{UDPOS} & 5.4 & 1.7 & 1.1 & 22.4 & 2.8 & 3.1 & 9.5 & 2.7 & 11.3 & N/A & N/A & 4.8 & 0.5 & 0.8 & 5.5 \\ \hline
\end{tabular}
\caption{Size of the test sets for the datasets in thousands of sentence pairs and sentences respectively. We use a 512 example subset of the released development sets to compute Shapley Values in all languages for all datasets.}
\label{descriptive}
\end{center}
\end{table*}

\subsection{Importance-Based Structured Pruning}\label{pruningMethod}
By using Shapley Values as the basis of structured pruning for multilingual tasks, we measure their practical ability to remove interference and generalize to unseen test data. Any signed importance metric can be used in this multilingual pruning process to test the metrics' ability to identify interference.

Our hypothesis is that attention heads with negative Shapley Values introduce interference broadly. Our pruning method reflects this by using the sign of our approximation directly. 
Using the empirical Bernstein inequality from Equation~\ref{bernEq}, we compute the upper bound Shapley Value. We then remove all attention heads for which this upper bound is negative, i.e. the set of attention heads which have probability $1-\delta$ of having negative Shapley Values. This is a parameter-free approach for deciding the number of heads to preserve. This approach is stable, with same set of negative heads is identified for pruning across 3 separate runs.

Alternatively, once Shapley Values are computed the model could be pruned to any sparsity level. Unlike prior pruning approaches besides \citet{michel-16}, we do not perform any weight updates following or during pruning and leave all parameters fixed. This provides constant time pruning to the desired size. We evaluate performance in this configurable pruning setting in \ref{configurablePruning}.

\section{Experiments}
\subsection{Datasets}
We evaluate our methodology on the Cross Lingual Natural Language Inference (XNLI) and Universal Dependencies Part-Of-Speech (UDPOS) tasks. These allow us to analyze the applicability of Attention Head Shapley Values to both sequence classification and structured prediction. We provide a description of dataset sizes in Table \ref{descriptive}.

\paragraph{Cross-Lingual Natural Language Inference (XNLI)}
We use the Cross Lingual Natural Language Inference Benchmark~\citep{conneau-etal-2018-xnli}. This dataset is aligned which allows us to control for possible confounding variation from the underlying semantics of the content. Given a premise and a hypothesis and tasks, XNLI is the task of classifying whether the hypothesis is entailed by the premise, contradicted by the premise, or neither. 

\paragraph{Universal Dependencies Part-of-Speech (UDPOS)}
For structured prediction, we evaluate on the Part-of-Speech tags from the Universal Dependencies v2 corpus~\citep{udpos}. The XTREME benchmark~\citep{xtreme} hypothesizes that structured prediction tasks are more language specific, with UDPOS having the largest cross-lingual gap in the benchmark. 

For direct comparison with our experiments on XNLI, we only retain the 13 languages from UDPOS have a development and test split and also exist in XNLI. Unlike XNLI, the subsections vary in data size and are not aligned across languages.

\subsection{Experimental Setup}
As the basis for our experiments, we finetune XLM-R Base~\citep{xlmr} 
using the Transformers library on only English data. Evaluation is done using the Datasets library~\citep{lhoest-etal-2021-datasets} implementation of the accuracy metrics. Finetuning and Shapley Value computation were both done on a single NVIDIA GeForce 12GB RTX 2080 Ti. We finetune following hyper-parameter tuning procedures from prior work: using \citet{xtreme} for XNLI and \citet{de-vries-etal-2022-make} for UDPOS. 

For all tasks and languages, we use the accuracy on 512 examples of the development set as the characteristic function for our coalitional game. Our pruning baselines include the gradient-based importance metric of \citet{michel-16} and the average of 10 randomly pruned networks. We prune the same number of heads pruned by our method for all strategies, since our baselines require selecting the number of heads to prune.

\subsection{Language Affinity}\label{affinity}
\begin{figure}
    \centering
    \includegraphics[width=0.48\textwidth]{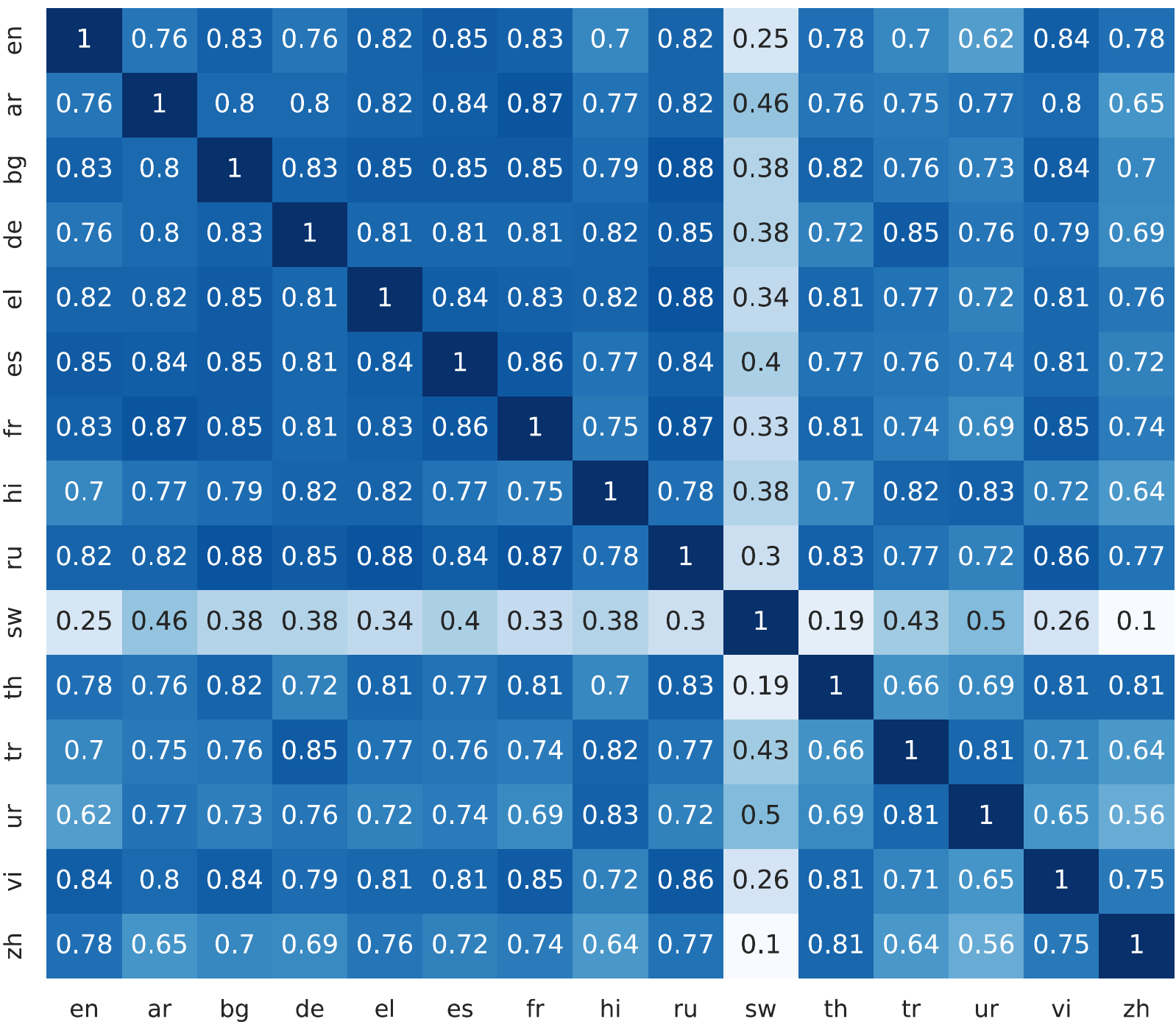}
    \caption{Spearman $\rho$ of Attention Head Shapley Values across languages in XNLI using XLM-R finetuned on the English training split.}
    \label{correlation}
\end{figure}
First, we analyze the Attention Head Shapley values for XNLI. We focus only on the role of the source language by using an aligned sample from XNLI to control our results for differences independent from language variation. In Figure \ref{figure2}, we visualize the results across English, Chinese, and Swahili. As expected from prior work~\citep{michel-16, voita-etal-2019-analyzing}, many heads have low magnitude Shapley Values indicating that they play no significant positive or negative role in the final network. We compare the similarity of Shapley Values learned across languages using Spearman's $\rho$ in Figure \ref{correlation} and find that Shapley Values are heavily correlated between all languages but Swahili, which is a major outlier. This cross-lingual consistency shows the stability of our learned values even across languages, in contrast with the instability shown by \citet{prasanna-etal-2020-bert}.

Despite this consistency, we find some attention heads demonstrate high language-specificity. Most notably, the fifth attention head in layer six is positive for Swahili, but strongly negative for all other 14 languages. This indicates that this head serves a function specific to Swahili within the model. We investigate the behavior of language-specific and language-agnostic heads further in Section \ref{analysis}.

\begin{table*}[!ht]
\begin{center}
\small	
\setlength\tabcolsep{1.5pt}
\renewcommand{\arraystretch}{1.2}
\begin{tabular}{c|ccccccccccccccc|}
\cline{2-16}
\multicolumn{1}{l|}{} & \multicolumn{15}{c|}{XNLI Accuracy} \\ \hline
\multicolumn{1}{|c|}{Pruning Strategy} & EN & AR & BG & DE & EL & ES & FR & HI & RU & SW & TH & TR & UR & VI & ZH \\ \hline
\multicolumn{1}{|c|}{No Pruning} & 84.1 & 70.6 & 76.7 & 76.8 & 75.4 & 79.8 & 77.7 & 70.0 & 74.7 & 63.4 & 70.6 & 71.9 & 65.9 & 73.3 & 73.5 \\
\multicolumn{1}{|c|}{Random} & 81.5$^-$ & 67.2$^-$ & 72.7$^-$ & 72.7$^-$ & 71.3$^-$ & 75.5$^-$ & 73.0$^-$ & 66.3$^-$ & 70.5$^-$ & 63.5 & 67.4$^-$ & 68.4$^-$ & 61.6$^-$ & 69.7$^-$ & 70.8$^-$ \\
\multicolumn{1}{|c|}{michel-16} & 84.3 & 71.0 & 77.3 & 77.4 & 72.8$^-$ & 80.2 & 78.4 & 71.5$^+$ & 75.2 & 63.1 & 70.7 & 71.7 & 66.9$^+$ & 73.3 & \textbf{77.2$^+$} \\ \hline
\multicolumn{1}{|c|}{Shapley Value ($\varphi_i$)} & \textbf{85.1$^+$} & \textbf{72.0$^+$} & \textbf{77.8$^+$} & \textbf{78.3$^+$} & \textbf{76.3} & \textbf{80.6} & \textbf{79.7$^+$} & \textbf{71.5$^+$} & \textbf{76.5$^+$} & \textbf{63.8} & \textbf{73.3$^+$} & \textbf{73.2$^+$} & \textbf{67.6$^+$} & \textbf{75.3$^+$} & \textbf{77.2$^+$} \\ \hline
\multicolumn{1}{|c|}{Pruned Heads ($K$)} & 4 & 6 & 6 & 5 & 4 & 5 & 5 & 7 & 5 & 5 & 7 & 5 & 6 & 6 & 9 \\ \hline
\end{tabular}
\\
\setlength\tabcolsep{2.85pt}
\begin{tabular}{c|ccccccccccccccc|}
\cline{2-16}
 & \multicolumn{15}{c|}{UDPOS Accuracy} \\ \hline
\multicolumn{1}{|c|}{Pruning Strategy} & EN & AR & BG & DE & EL & ES & FR & HI & RU & SW & TH & TR & UR & VI & ZH \\ \hline
\multicolumn{1}{|c|}{No Pruning} & \textbf{95.7} & 75.1 & \textbf{90.9} & \textbf{88.8} & 71.5 & \textbf{89.8} & 81.3 & 73.9 & \textbf{88.2} & - & - & \textbf{78.7} & 67.3 & \textbf{66.3} & 50.2 \\
\multicolumn{1}{|c|}{Random} & \textbf{95.7} & 74.3$^-$ & \textbf{90.9} & \textbf{88.8} & 71.8 & \textbf{89.8} & 81.4 & 73.7 & \textbf{88.2} & - & - & \textbf{78.7} & 67.5 & \textbf{66.3} & 55.3$^+$ \\
\multicolumn{1}{|c|}{\citet{michel-16}} & \textbf{95.7} & 75.1 & \textbf{90.9} & \textbf{88.8} & 71.1 & \textbf{89.8} & 81.1 & 73.8 & \textbf{88.2} & - & - & \textbf{78.7} & 67.3 & \textbf{66.3} & 48.9$^-$ \\ \hline
\multicolumn{1}{|c|}{Shapley Value ($\varphi_i$)} & \textbf{95.7} & \textbf{76.6}$^+$ & \textbf{90.9} & \textbf{88.8} & \textbf{72.8}$^+$ & \textbf{89.8} & \textbf{82.6}$^+$ & \textbf{75.6}$^+$ & \textbf{88.2} & - & - & \textbf{78.7} & \textbf{69.5}$^+$ & \textbf{66.3} & \textbf{62.6}$^+$ \\ \hline
\multicolumn{1}{|c|}{Pruned Heads ($K$)} & 0 & 4 & 0 & 0 & 4 & 0 & 2 & 2 & 0 & - & - & 0 & 4 & 0 & 18 \\ \hline
\end{tabular}
\end{center}
\caption{Accuracy for UDPOS and XNLI after pruning according to importance metrics. For all metrics, we remove the Bottom-$K$ heads ($K=\lvert\{H_i \mid \varphi_i < 0\}\rvert$) according to that metric. $^+$ and $^-$ indicate significant ($P<0.05$) improvement and harm by a pairwise bootstrap test. Model parameters remain fixed for all methods.}\label{table1}
\end{table*}

It is worth noting that the outlier, Swahili, is the language with the fewest number of examples in the data used in the pretraining of XLM-R. Whether the large variation between Swahili and all other languages is induced by linguistic features or the training dynamics of low-resource languages within multilingual models is unclear. We leave this to be explored further in future work.

\subsection{Targeted Pruning}
To understand the practical applicability of the resulting Shapley Values, we evaluate models before and after pruning all attention heads with negative Shapley Values as described in Section \ref{pruningMethod}. 

Each resulting language-specific model can be represented with only the 144 mask parameters which indicate whether each attention head is removed or kept. Therefore, this pruning can be seen alternatively as a parameter-efficient learning method, using $5*10^{-7}\%$ of the parameters it would require to finetune the model for each language. 

\paragraph{XNLI} 
\noindent In Table \ref{table1}, we report the accuracy of models after targeted pruning across all languages for both XNLI and UDPOS. For XNLI, we see that targeted pruning improves performance by an average of +1.59 across all 15 languages with the maximum improvement being in Chinese (+3.78) and the minimum improvement in Swahili (+0.37). While it is expected that languages more closely related to English would benefit less from pruning, even closely related languages such as French (+1.97) and German (+1.53) are improved significantly.

\paragraph{UDPOS} 
\noindent Improvements in UDPOS vary to a higher degree. Only 6 out of 13 languages improve after pruning, with the rest identical with no negative Shapley Values. The largest improvement is again in Chinese (+12.4) and the smallest in French (+1.3). In the case of Chinese, this is a 24.7\% improvement purely by removing attention heads. Across the languages which were pruned, the average improvement is 3.4 -- reducing the cross-lingual gap~\citep{xtreme} by 0.7. 

\paragraph{Comparison to Baselines}
\noindent Randomly pruning is ineffectual or harms performance in both tasks, indicating that pruning alone is not the source of our improvement. Pruning according to the gradient-based metric proposed by \citet{michel-16} maintains rather than improves performance. This supports our hypothesis that methods which use the magnitude of gradients largely identify non-impactful heads as opposed to harmful heads.

\begin{table*}[!ht]
\begin{center}
\small	
\setlength\tabcolsep{1.5pt}
\renewcommand{\arraystretch}{1.2}
\begin{tabular}{|c|ccccccccccccccc|}
\hline
Pruning Strategy            & EN                & AR                & BG                & DE                & EL            & ES            & FR                & HI                & RU                & SW            & TH                & TR                & UR                & VI                & ZH                \\ \hline
No Pruning                  & 84.1              & 70.6              & 76.7              & 76.8              & 75.4 & 79.8 & 77.7              & 70.0              & 74.7              & \textbf{63.4} & 70.6              & 71.9              & 65.9              & 73.3              & 73.5              \\
Random                      & 81.7$^-$          & 67.1$^-$          & 72.3$^-$          & 72.9$^-$          & 71.1$^-$      & 75.1$^-$      & 73.5$^-$          & 65.7$^-$          & 71$^-$            & 60.7$^-$      & 67$^-$            & 68.3$^-$          & 61$^-$            & 69.7$^-$          & 70.7$^-$          \\
\citet{michel-16}           & 84.3              & 70.3              & 76.7              & 77.1              & 75.9          & 80.1          & 77.9              & 70.1              & 75.1              & 62.9          & 71.6              & 72.5              & 66.1              & 74.7$^+$              & 74.5$^+$              \\ \hline
Shapley Value ($\varphi_i$) & \textbf{85.1$^+$} & \textbf{72.0$^+$} & \textbf{77.8$^+$} & \textbf{79.4$^+$} & \textbf{76.3} & \textbf{80.6} & \textbf{79.7$^+$} & \textbf{71.5$^+$} & \textbf{76.5$^+$} & 63.3      & \textbf{73.1$^+$} & \textbf{73.1$^+$} & \textbf{68.4$^+$} & \textbf{75.2$^+$} & \textbf{76.3$^+$} \\ \hline
\end{tabular}
\end{center}
\caption{Accuracy for XNLI after pruning using importance metrics from English. For all metrics, we remove the Bottom-$K$ heads ($K=\lvert\{H_i \mid \varphi_i < 0\}\rvert$) according to that metric. $^+$ and $^-$ indicate significant ($P<0.05$) improvement and harm by a pairwise bootstrap test.}\label{table2}
\end{table*}
\subsection{Zero-Shot Pruning}\label{zero-shot}
Given the high rank correlation between many of the languages, we evaluate transferability by using the Shapley Values for English to prune the model for all languages. We report results in Table \ref{table2}.

\paragraph{XNLI} On XNLI, surprisingly, this transferred pruning across languages has similar benefits to our targeted pruning results despite only being learned for English. Two languages (Urdu and German) achieve better results in the zero-shot pruning than they did in the targeted pruning, five achieve worse results, and the remaining eight are equivalent. 

It is likely that the strength of zero-shot transfer is largely due to the removal of the fifth head of layer six, which is one of the top 2 most negative heads for all languages barring Swahili. Interestingly, the Attention Head Shapley Values for Swahili also have the lowest rank correlation with English of any language.

\paragraph{UDPOS} However, UDPOS highlights the major shortcoming of zero-shot pruning: all attention heads receive a positive Shapley Value for English for UDPOS. This means that no zero-shot pruning is performed even though targeted pruning despite benefits for other languages shown in Table \ref{table1}.

\subsection{Iterative Pruning of Attention Heads}
\begin{figure}[t]
\begin{center}
\includegraphics[width=0.48\textwidth]{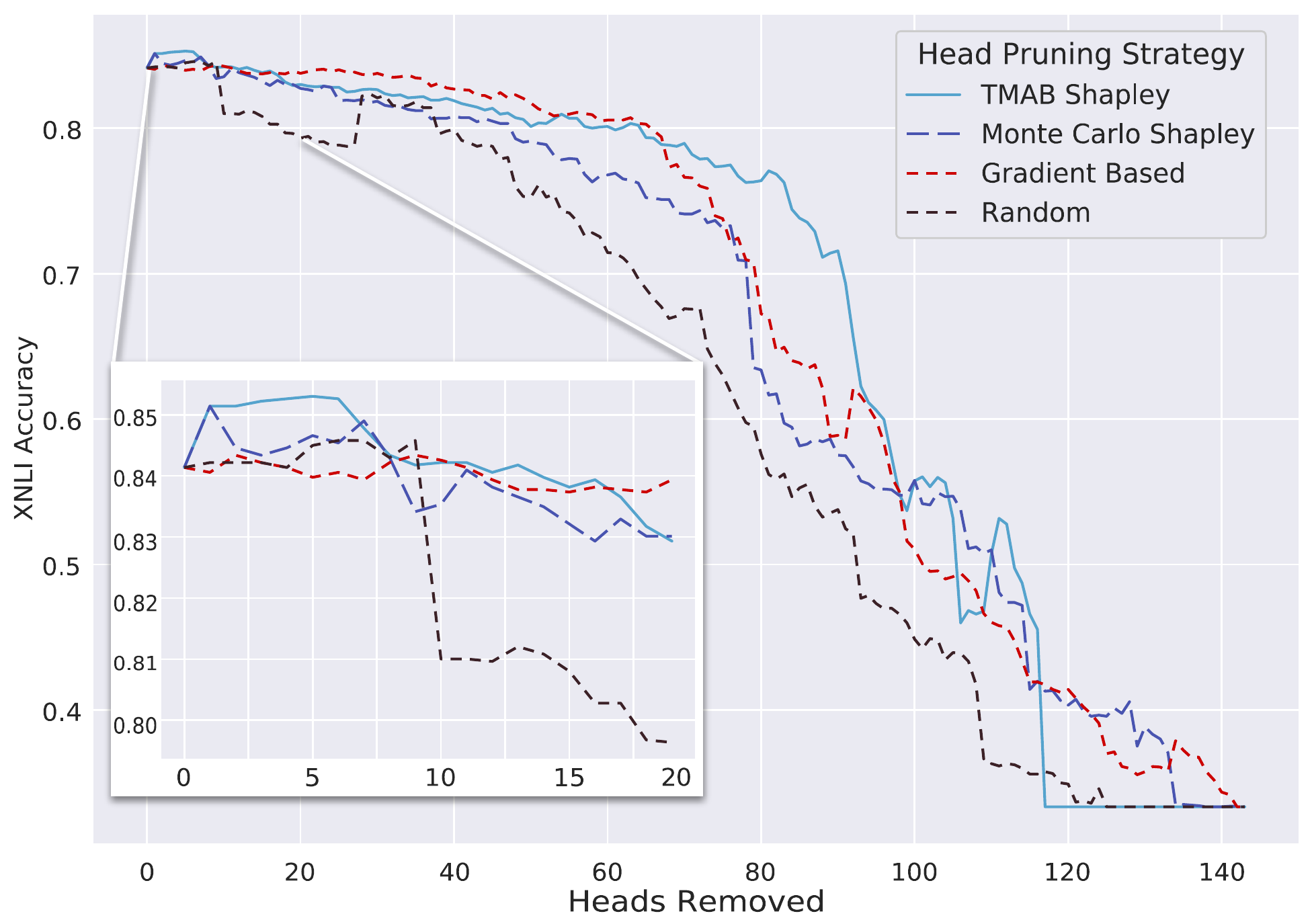}
\caption{Evolution of XNLI Accuracy as Heads are removed according to different pruning strategies.}\label{configurable}
\end{center}
\end{figure}

Finally, we evaluate the effectiveness of Shapley Values as a ranking methodology for the iterative pruning evaluation performed by \citet{michel-16}. Iterative pruning evaluates how well each importance ranking captures the combinatorial effects of removing attention heads at different compute budgets. We compare random pruning, the gradient-based approach from \citet{michel-16}, and Shapley Values computed through plain Monte Carlo simulation and Shapley Values using Truncation and Multi-Armed Bandit optimization (TMAB). We plot results in Figure \ref{configurable}.

Averaged across all levels of sparsity, our method outperforms the Random baseline (+5.8), Monte Carlo Shapley Values (+1.6), and the Gradient baseline (+0.6). Our method is the most effective at identifying strongly harmful heads in early stages of pruning with performance improving compared to the unpruned model for the first 6 heads removed. Additionally, our method is superior at pruning the model to approximately half of its original size, achieving the largest performance gap at 44\% of model capacity outperforming the Gradient baseline, Monte Carlo Shapley Values, and the Random Baseline by +12.2, +15.1, and +20.9 respectively. 

It is worth noting that the gradient baseline outperforms our method at very high sparsity when more than 80\% of heads are pruned. However, at this sparsity neither method results in a model which performs well above chance. 
\label{configurablePruning}

\begin{figure*}[!t]
\begin{center}
\includegraphics[width=\paperwidth-6cm]{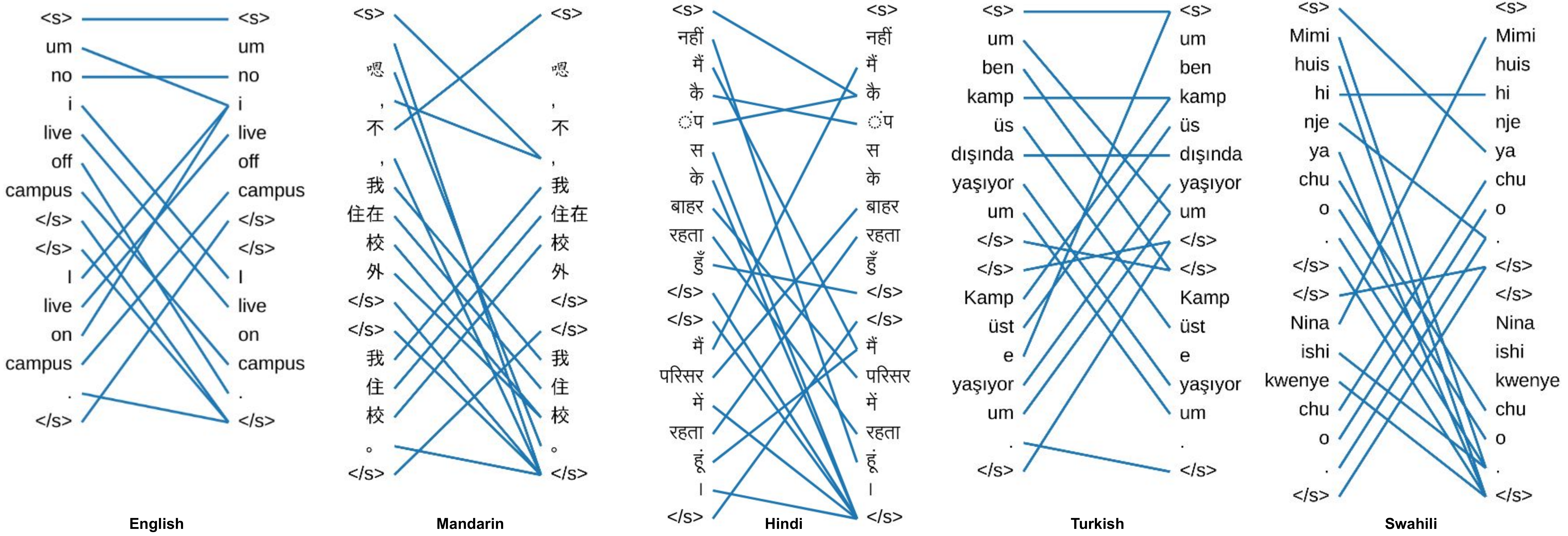}
\vspace{-5pt}
\caption{Attention of Layer 4, Head 8 of our XNLI model which is identified as language-agnostic. For clarity, we connect the left token to the token on the right which receives the largest attention weight.}\label{languageagnostic}
\end{center}
\end{figure*}
\begin{figure*}[!t]
\begin{center}
\includegraphics[width=\paperwidth-6cm]{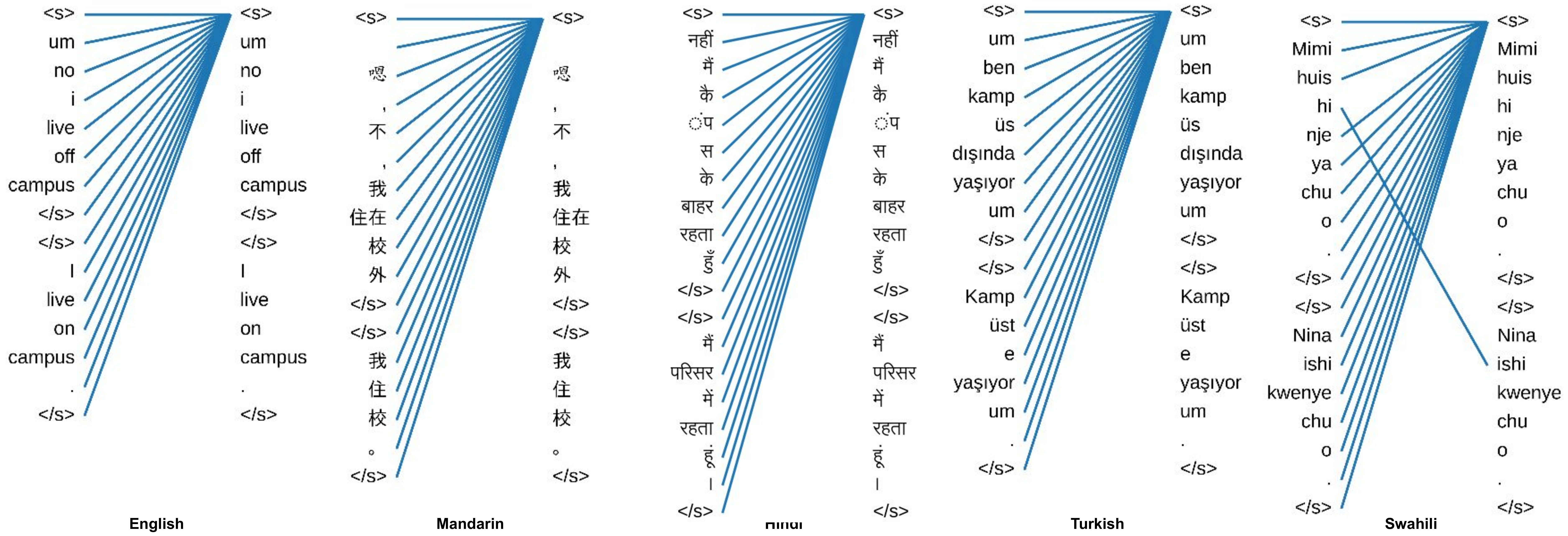}
\vspace{-5pt}
\caption{Attention of Layer 6, Head 5 of our XNLI model which is identified as language-specific.}\label{language-specific}
\end{center}
\end{figure*}
\section{Qualitative Attention Analysis}\label{analysis}
In order to provide intuition into the function of attention heads, prior work has turned to attention visualization as the basis for qualitative analysis of the inner workings of transformer models. \citet{clark-etal-2019-bert} and \citet{hoover-etal-2020-exbert} both find human interpretable patterns within attention heads.


We visualize the attention patterns of outlier attention heads using BertViz~\citep{vig-2019-multiscale} from our model to give a qualitative understanding of the attention head patterns associated with language-agnostic and language-specific heads.

\subsection{Language-Agnostic Heads}
To identify language-agnostic heads, we take the top 20 attention heads for each language and the intersection of these sets. In Figure \ref{languageagnostic}, we visualize the attention pattern of the higher ranked of the two heads which meet this criteria, though qualitatively both seem to have the same function. This head exhibits the same function across all languages: matching words from the premise to near synonyms in the hypothesis and vice versa.

This pattern is clearly applicable to NLI, which requires finding commonalities and contradictions across the premise and hypothesis. On the other hand, it does not require any knowledge of language-specific syntax or morphology since token semantics combined with the separator tokens is sufficient to connect synonyms across sentences.

\subsection{Language-Specific Heads}
As highlighted in Section \ref{affinity}, the fifth head of layer six has a positive Shapley Value only for Swahili. In Figure \ref{language-specific}, we see that this head exhibits unique behavior for Swahili, connecting the suffixes of "\emph{Mimi Huishi}" and "\emph{Ninaishi}" meaning "\emph{I live}" in the Habitual and Present tense respectively.

Beyond a single example, the attention pattern of this head varies quantitatively for Swahili. Using \citet{clark-etal-2019-bert} hypothesis that attention to separator tokens indicates an inapplicable learned pattern, we look at the percentage of sentences where all tokens attend primarily to separators. This criterion is true in 56\% Swahili XNLI inputs, but only 41\% of non-Swahili inputs on average ($\sigma = 4.3\%$).

Given the small negative performance impact removing negative English heads has on Swahili in \ref{zero-shot}, we hypothesize that this head captures an infrequent pattern in Swahili, but more commonly introduces noise to other languages. However, systematic analysis of structural language affinity is a promising area for further work.

\section{Conclusions \& Future Work}
In this work, we developed a simple yet effective approach to identify language-specific structural components of multilingual transformer language models by leveraging Shapley Values. We demonstrated that the resulting values do exhibit language affinity, varying across languages. We then applied these Attention Head Shapley Values to improve cross-lingual performance through pruning for both sequence classification and structured prediction. Finally, we performed attention visualization and provided insights on language-agnostic and language-specific attention heads. 

Future work should attempt to understand the relationships between linguistic features, training data volume, and the language-specificity of attention heads more systematically. Additionally, the benefits of removing heads motivates work which reduces cross-lingual interference introduced by language-specific components during pre-training, such as pruning during pretraining or utilizing sparsely activated networks.

\section{Limitations}
Even with our computational optimizations, using Shapley Values as an importance metric requires a significant computational cost compared to gradient-based methods: gradient-based methods take approximately 3.33e14 FLOPs and our optimized Shapley Computation takes approximately 3.27e16 FLOPs. While the computation is parallelizable, it took several days on a single GPU to compute accurate estimates even for small validation sets. The computational expense is reasonable for understanding the behavior of base models more deeply, but limits the use of this method as a rapid iteration tool.
Additionally, we rely on analysis of attention patterns to help ground our findings. However, there is debate as to whether analysis of attention patterns is a sound analytical tool~\citep{attention-is-not, attention-is-not-not}.

\section{Acknowledgements}
We are thankful to Chris Hidey, Yanzhe Zhang, Hongxin Zhang, Caleb Ziems and our anonymous reviewers for their feedback. This work is supported in part by Cisco, an Amazon Faculty Research Award and NSF grant IIS-2144562.
\bibliographystyle{acl_natbib}
\bibliography{acl2021}

\begin{thebibliography}{47}
\expandafter\ifx\csname natexlab\endcsname\relax\def\natexlab#1{#1}\fi

\bibitem[{Ansell et~al.(2022)Ansell, Ponti, Korhonen, and
  Vuli{\'c}}]{ansell-etal-2022-composable}
Alan Ansell, Edoardo Ponti, Anna Korhonen, and Ivan Vuli{\'c}. 2022.
\newblock \href {https://doi.org/10.18653/v1/2022.acl-long.125} {Composable
  sparse fine-tuning for cross-lingual transfer}.
\newblock In \emph{Proceedings of the 60th Annual Meeting of the Association
  for Computational Linguistics (Volume 1: Long Papers)}, pages 1778--1796,
  Dublin, Ireland. Association for Computational Linguistics.

\bibitem[{Ansell et~al.(2021)Ansell, Ponti, Pfeiffer, Ruder, Glava{\v{s}},
  Vuli{\'c}, and Korhonen}]{ansell-etal-2021-mad-g}
Alan Ansell, Edoardo~Maria Ponti, Jonas Pfeiffer, Sebastian Ruder, Goran
  Glava{\v{s}}, Ivan Vuli{\'c}, and Anna Korhonen. 2021.
\newblock \href {https://doi.org/10.18653/v1/2021.findings-emnlp.410}
  {{MAD}-{G}: {M}ultilingual adapter generation for efficient cross-lingual
  transfer}.
\newblock In \emph{Findings of the Association for Computational Linguistics:
  EMNLP 2021}, pages 4762--4781, Punta Cana, Dominican Republic. Association
  for Computational Linguistics.

\bibitem[{Artetxe et~al.(2018)Artetxe, Labaka, and
  Agirre}]{artetxe-etal-2018-robust}
Mikel Artetxe, Gorka Labaka, and Eneko Agirre. 2018.
\newblock \href {https://doi.org/10.18653/v1/P18-1073} {A robust self-learning
  method for fully unsupervised cross-lingual mappings of word embeddings}.
\newblock In \emph{Proceedings of the 56th Annual Meeting of the Association
  for Computational Linguistics (Volume 1: Long Papers)}, pages 789--798,
  Melbourne, Australia. Association for Computational Linguistics.

\bibitem[{Artetxe et~al.(2020)Artetxe, Ruder, and
  Yogatama}]{artetxe-etal-2020-cross}
Mikel Artetxe, Sebastian Ruder, and Dani Yogatama. 2020.
\newblock \href {https://doi.org/10.18653/v1/2020.acl-main.421} {On the
  cross-lingual transferability of monolingual representations}.
\newblock In \emph{Proceedings of the 58th Annual Meeting of the Association
  for Computational Linguistics}, pages 4623--4637, Online. Association for
  Computational Linguistics.

\bibitem[{Artetxe and Schwenk(2019)}]{artetxe-schwenk-2019-massively}
Mikel Artetxe and Holger Schwenk. 2019.
\newblock \href {https://doi.org/10.1162/tacl_a_00288} {Massively multilingual
  sentence embeddings for zero-shot cross-lingual transfer and beyond}.
\newblock \emph{Transactions of the Association for Computational Linguistics},
  7:597--610.

\bibitem[{Castro et~al.(2009)Castro, Gómez, and Tejada}]{monte}
Javier Castro, Daniel Gómez, and Juan Tejada. 2009.
\newblock \href {https://doi.org/https://doi.org/10.1016/j.cor.2008.04.004}
  {Polynomial calculation of the shapley value based on sampling}.
\newblock \emph{Computers \& Operations Research}, 36(5):1726--1730.
\newblock Selected papers presented at the Tenth International Symposium on
  Locational Decisions (ISOLDE X).

\bibitem[{Chen et~al.(2020)Chen, Frankle, Chang, Liu, Zhang, Wang, and
  Carbin}]{lotterytransformers}
Tianlong Chen, Jonathan Frankle, Shiyu Chang, Sijia Liu, Yang Zhang, Zhangyang
  Wang, and Michael Carbin. 2020.
\newblock \href
  {https://proceedings.neurips.cc/paper/2020/file/b6af2c9703f203a2794be03d443af2e3-Paper.pdf}
  {The lottery ticket hypothesis for pre-trained bert networks}.
\newblock In \emph{Advances in Neural Information Processing Systems},
  volume~33, pages 15834--15846. Curran Associates, Inc.

\bibitem[{Clark et~al.(2019)Clark, Khandelwal, Levy, and
  Manning}]{clark-etal-2019-bert}
Kevin Clark, Urvashi Khandelwal, Omer Levy, and Christopher~D. Manning. 2019.
\newblock \href {https://doi.org/10.18653/v1/W19-4828} {What does {BERT} look
  at? an analysis of {BERT}{'}s attention}.
\newblock In \emph{Proceedings of the 2019 ACL Workshop BlackboxNLP: Analyzing
  and Interpreting Neural Networks for NLP}, pages 276--286, Florence, Italy.
  Association for Computational Linguistics.

\bibitem[{Conneau et~al.(2019)Conneau, Khandelwal, Goyal, Chaudhary, Wenzek,
  Guzmán, Grave, Ott, Zettlemoyer, and Stoyanov}]{xlmr}
Alexis Conneau, Kartikay Khandelwal, Naman Goyal, Vishrav Chaudhary, Guillaume
  Wenzek, Francisco Guzmán, Edouard Grave, Myle Ott, Luke Zettlemoyer, and
  Veselin Stoyanov. 2019.
\newblock \href {https://doi.org/10.48550/ARXIV.1911.02116} {Unsupervised
  cross-lingual representation learning at scale}.

\bibitem[{Conneau et~al.(2018)Conneau, Rinott, Lample, Williams, Bowman,
  Schwenk, and Stoyanov}]{conneau-etal-2018-xnli}
Alexis Conneau, Ruty Rinott, Guillaume Lample, Adina Williams, Samuel Bowman,
  Holger Schwenk, and Veselin Stoyanov. 2018.
\newblock \href {https://doi.org/10.18653/v1/D18-1269} {{XNLI}: Evaluating
  cross-lingual sentence representations}.
\newblock In \emph{Proceedings of the 2018 Conference on Empirical Methods in
  Natural Language Processing}, pages 2475--2485, Brussels, Belgium.
  Association for Computational Linguistics.

\bibitem[{Conneau et~al.(2020)Conneau, Wu, Li, Zettlemoyer, and
  Stoyanov}]{conneau-etal-2020-emerging}
Alexis Conneau, Shijie Wu, Haoran Li, Luke Zettlemoyer, and Veselin Stoyanov.
  2020.
\newblock \href {https://doi.org/10.18653/v1/2020.acl-main.536} {Emerging
  cross-lingual structure in pretrained language models}.
\newblock In \emph{Proceedings of the 58th Annual Meeting of the Association
  for Computational Linguistics}, pages 6022--6034, Online. Association for
  Computational Linguistics.

\bibitem[{de~Vries et~al.(2022)de~Vries, Wieling, and
  Nissim}]{de-vries-etal-2022-make}
Wietse de~Vries, Martijn Wieling, and Malvina Nissim. 2022.
\newblock \href {https://doi.org/10.18653/v1/2022.acl-long.529} {Make the best
  of cross-lingual transfer: Evidence from {POS} tagging with over 100
  languages}.
\newblock In \emph{Proceedings of the 60th Annual Meeting of the Association
  for Computational Linguistics (Volume 1: Long Papers)}, pages 7676--7685,
  Dublin, Ireland. Association for Computational Linguistics.

\bibitem[{Frankle and Carbin(2019)}]{lotteryticket}
Jonathan Frankle and Michael Carbin. 2019.
\newblock \href {https://openreview.net/forum?id=rJl-b3RcF7} {The lottery
  ticket hypothesis: Finding sparse, trainable neural networks}.
\newblock In \emph{7th International Conference on Learning Representations,
  {ICLR} 2019, New Orleans, LA, USA, May 6-9, 2019}. OpenReview.net.

\bibitem[{Ghorbani and Zou(2019)}]{datashap}
Amirata Ghorbani and James Zou. 2019.
\newblock \href {https://proceedings.mlr.press/v97/ghorbani19c.html} {Data
  shapley: Equitable valuation of data for machine learning}.
\newblock In \emph{Proceedings of the 36th International Conference on Machine
  Learning}, volume~97 of \emph{Proceedings of Machine Learning Research},
  pages 2242--2251. PMLR.

\bibitem[{Ghorbani and Zou(2020)}]{neuronshap}
Amirata Ghorbani and James~Y Zou. 2020.
\newblock \href
  {https://proceedings.neurips.cc/paper/2020/file/41c542dfe6e4fc3deb251d64cf6ed2e4-Paper.pdf}
  {Neuron shapley: Discovering the responsible neurons}.
\newblock In \emph{Advances in Neural Information Processing Systems},
  volume~33, pages 5922--5932. Curran Associates, Inc.

\bibitem[{Hoover et~al.(2020)Hoover, Strobelt, and
  Gehrmann}]{hoover-etal-2020-exbert}
Benjamin Hoover, Hendrik Strobelt, and Sebastian Gehrmann. 2020.
\newblock \href {https://doi.org/10.18653/v1/2020.acl-demos.22} {ex{BERT}: {A}
  {V}isual {A}nalysis {T}ool to {E}xplore {L}earned {R}epresentations in
  {T}ransformer {M}odels}.
\newblock In \emph{Proceedings of the 58th Annual Meeting of the Association
  for Computational Linguistics: System Demonstrations}, pages 187--196,
  Online. Association for Computational Linguistics.

\bibitem[{Hu et~al.(2021)Hu, Johnson, Firat, Siddhant, and
  Neubig}]{hu-etal-2021-explicit}
Junjie Hu, Melvin Johnson, Orhan Firat, Aditya Siddhant, and Graham Neubig.
  2021.
\newblock \href {https://doi.org/10.18653/v1/2021.naacl-main.284} {Explicit
  alignment objectives for multilingual bidirectional encoders}.
\newblock In \emph{Proceedings of the 2021 Conference of the North American
  Chapter of the Association for Computational Linguistics: Human Language
  Technologies}, pages 3633--3643, Online. Association for Computational
  Linguistics.

\bibitem[{Hu et~al.(2020)Hu, Ruder, Siddhant, Neubig, Firat, and
  Johnson}]{xtreme}
Junjie Hu, Sebastian Ruder, Aditya Siddhant, Graham Neubig, Orhan Firat, and
  Melvin Johnson. 2020.
\newblock \href {http://arxiv.org/abs/2003.11080} {Xtreme: A massively
  multilingual multi-task benchmark for evaluating cross-lingual
  generalization}.
\newblock \emph{CoRR}, abs/2003.11080.

\bibitem[{Jain and Wallace(2019)}]{attention-is-not}
Sarthak Jain and Byron~C. Wallace. 2019.
\newblock \href {https://doi.org/10.18653/v1/N19-1357} {{A}ttention is not
  {E}xplanation}.
\newblock In \emph{Proceedings of the 2019 Conference of the North {A}merican
  Chapter of the Association for Computational Linguistics: Human Language
  Technologies, Volume 1 (Long and Short Papers)}, pages 3543--3556,
  Minneapolis, Minnesota. Association for Computational Linguistics.

\bibitem[{Joulin et~al.(2018)Joulin, Bojanowski, Mikolov, J{\'e}gou, and
  Grave}]{joulin-etal-2018-loss}
Armand Joulin, Piotr Bojanowski, Tomas Mikolov, Herv{\'e} J{\'e}gou, and
  Edouard Grave. 2018.
\newblock \href {https://doi.org/10.18653/v1/D18-1330} {Loss in translation:
  Learning bilingual word mapping with a retrieval criterion}.
\newblock In \emph{Proceedings of the 2018 Conference on Empirical Methods in
  Natural Language Processing}, pages 2979--2984, Brussels, Belgium.
  Association for Computational Linguistics.

\bibitem[{Kovaleva et~al.(2019)Kovaleva, Romanov, Rogers, and
  Rumshisky}]{kovaleva-etal-2019-revealing}
Olga Kovaleva, Alexey Romanov, Anna Rogers, and Anna Rumshisky. 2019.
\newblock \href {https://doi.org/10.18653/v1/D19-1445} {Revealing the dark
  secrets of {BERT}}.
\newblock In \emph{Proceedings of the 2019 Conference on Empirical Methods in
  Natural Language Processing and the 9th International Joint Conference on
  Natural Language Processing (EMNLP-IJCNLP)}, pages 4365--4374, Hong Kong,
  China. Association for Computational Linguistics.

\bibitem[{Lee et~al.(2019)Lee, Ajanthan, and Torr}]{lee2018snip}
Namhoon Lee, Thalaiyasingam Ajanthan, and Philip Torr. 2019.
\newblock \href {https://openreview.net/forum?id=B1VZqjAcYX} {{SNIP}:
  Single-shot network pruning based on connection sensitivity}.
\newblock In \emph{International Conference on Learning Representations}.

\bibitem[{Lhoest et~al.(2021)Lhoest, Villanova~del Moral, Jernite, Thakur, von
  Platen, Patil, Chaumond, Drame, Plu, Tunstall, Davison, {\v{S}}a{\v{s}}ko,
  Chhablani, Malik, Brandeis, Le~Scao, Sanh, Xu, Patry, McMillan-Major, Schmid,
  Gugger, Delangue, Matussi{\`e}re, Debut, Bekman, Cistac, Goehringer, Mustar,
  Lagunas, Rush, and Wolf}]{lhoest-etal-2021-datasets}
Quentin Lhoest, Albert Villanova~del Moral, Yacine Jernite, Abhishek Thakur,
  Patrick von Platen, Suraj Patil, Julien Chaumond, Mariama Drame, Julien Plu,
  Lewis Tunstall, Joe Davison, Mario {\v{S}}a{\v{s}}ko, Gunjan Chhablani,
  Bhavitvya Malik, Simon Brandeis, Teven Le~Scao, Victor Sanh, Canwen Xu,
  Nicolas Patry, Angelina McMillan-Major, Philipp Schmid, Sylvain Gugger,
  Cl{\'e}ment Delangue, Th{\'e}o Matussi{\`e}re, Lysandre Debut, Stas Bekman,
  Pierric Cistac, Thibault Goehringer, Victor Mustar, Fran{\c{c}}ois Lagunas,
  Alexander Rush, and Thomas Wolf. 2021.
\newblock \href {https://doi.org/10.18653/v1/2021.emnlp-demo.21} {Datasets: A
  community library for natural language processing}.
\newblock In \emph{Proceedings of the 2021 Conference on Empirical Methods in
  Natural Language Processing: System Demonstrations}, pages 175--184, Online
  and Punta Cana, Dominican Republic. Association for Computational
  Linguistics.

\bibitem[{Liu et~al.(2020)Liu, Gu, Goyal, Li, Edunov, Ghazvininejad, Lewis, and
  Zettlemoyer}]{mbart}
Yinhan Liu, Jiatao Gu, Naman Goyal, Xian Li, Sergey Edunov, Marjan
  Ghazvininejad, Mike Lewis, and Luke Zettlemoyer. 2020.
\newblock \href {https://doi.org/10.1162/tacl_a_00343} {Multilingual denoising
  pre-training for neural machine translation}.
\newblock \emph{Transactions of the Association for Computational Linguistics},
  8:726--742.

\bibitem[{Lundberg and Lee(2017)}]{shap}
Scott~M. Lundberg and Su-In Lee. 2017.
\newblock A unified approach to interpreting model predictions.
\newblock NIPS'17, page 4768–4777, Red Hook, NY, USA. Curran Associates Inc.

\bibitem[{Ma et~al.(2021)Ma, Zhang, Lou, Wang, and
  Vosoughi}]{ma-etal-2021-contributions}
Weicheng Ma, Kai Zhang, Renze Lou, Lili Wang, and Soroush Vosoughi. 2021.
\newblock \href {https://doi.org/10.18653/v1/2021.acl-long.152} {Contributions
  of transformer attention heads in multi- and cross-lingual tasks}.
\newblock In \emph{Proceedings of the 59th Annual Meeting of the Association
  for Computational Linguistics and the 11th International Joint Conference on
  Natural Language Processing (Volume 1: Long Papers)}, pages 1956--1966,
  Online. Association for Computational Linguistics.

\bibitem[{Mann and Yarowsky(2001)}]{mann-yarowsky-2001-multipath}
Gideon~S. Mann and David Yarowsky. 2001.
\newblock \href {https://aclanthology.org/N01-1020} {Multipath translation
  lexicon induction via bridge languages}.
\newblock In \emph{Second Meeting of the North {A}merican Chapter of the
  Association for Computational Linguistics}.

\bibitem[{Maurer and Pontil(2009)}]{empiricalbern}
Andreas Maurer and Massimiliano Pontil. 2009.
\newblock \href {http://www.cs.mcgill.ca/\%7Ecolt2009/papers/012.pdf\#page=1}
  {Empirical bernstein bounds and sample-variance penalization}.
\newblock In \emph{{COLT} 2009 - The 22nd Conference on Learning Theory,
  Montreal, Quebec, Canada, June 18-21, 2009}.

\bibitem[{Mayhew et~al.(2017)Mayhew, Tsai, and Roth}]{mayhew-etal-2017-cheap}
Stephen Mayhew, Chen-Tse Tsai, and Dan Roth. 2017.
\newblock \href {https://doi.org/10.18653/v1/D17-1269} {Cheap translation for
  cross-lingual named entity recognition}.
\newblock In \emph{Proceedings of the 2017 Conference on Empirical Methods in
  Natural Language Processing}, pages 2536--2545, Copenhagen, Denmark.
  Association for Computational Linguistics.

\bibitem[{McDonald et~al.(2011)McDonald, Petrov, and Hall}]{mcdonald2011multi}
Ryan McDonald, Slav Petrov, and Keith Hall. 2011.
\newblock Multi-source transfer of delexicalized dependency parsers.
\newblock In \emph{Proceedings of the 2011 conference on empirical methods in
  natural language processing}, pages 62--72.

\bibitem[{Michel et~al.(2019)Michel, Levy, and Neubig}]{michel-16}
Paul Michel, Omer Levy, and Graham Neubig. 2019.
\newblock \href
  {https://proceedings.neurips.cc/paper/2019/file/2c601ad9d2ff9bc8b282670cdd54f69f-Paper.pdf}
  {Are sixteen heads really better than one?}
\newblock In \emph{Advances in Neural Information Processing Systems},
  volume~32. Curran Associates, Inc.

\bibitem[{Mnih et~al.(2008)Mnih, Szepesv\'{a}ri, and Audibert}]{bernStopping}
Volodymyr Mnih, Csaba Szepesv\'{a}ri, and Jean-Yves Audibert. 2008.
\newblock \href {https://doi.org/10.1145/1390156.1390241} {Empirical bernstein
  stopping}.
\newblock In \emph{Proceedings of the 25th International Conference on Machine
  Learning}, ICML '08, page 672–679, New York, NY, USA. Association for
  Computing Machinery.

\bibitem[{Nivre et~al.(2020)Nivre, de~Marneffe, Ginter, Haji{\v{c}}, Manning,
  Pyysalo, Schuster, Tyers, and Zeman}]{udpos}
Joakim Nivre, Marie-Catherine de~Marneffe, Filip Ginter, Jan Haji{\v{c}},
  Christopher~D. Manning, Sampo Pyysalo, Sebastian Schuster, Francis Tyers, and
  Daniel Zeman. 2020.
\newblock \href {https://aclanthology.org/2020.lrec-1.497} {{U}niversal
  {D}ependencies v2: An evergrowing multilingual treebank collection}.
\newblock In \emph{Proceedings of the 12th Language Resources and Evaluation
  Conference}, pages 4034--4043, Marseille, France. European Language Resources
  Association.

\bibitem[{Pfeiffer et~al.(2020)Pfeiffer, Vuli{\'c}, Gurevych, and
  Ruder}]{pfeiffer-etal-2020-mad}
Jonas Pfeiffer, Ivan Vuli{\'c}, Iryna Gurevych, and Sebastian Ruder. 2020.
\newblock \href {https://doi.org/10.18653/v1/2020.emnlp-main.617} {{MAD-X}:
  {A}n {A}dapter-{B}ased {F}ramework for {M}ulti-{T}ask {C}ross-{L}ingual
  {T}ransfer}.
\newblock In \emph{Proceedings of the 2020 Conference on Empirical Methods in
  Natural Language Processing (EMNLP)}, pages 7654--7673, Online. Association
  for Computational Linguistics.

\bibitem[{Pires et~al.(2019)Pires, Schlinger, and Garrette}]{mbert}
Telmo Pires, Eva Schlinger, and Dan Garrette. 2019.
\newblock \href {https://doi.org/10.18653/v1/P19-1493} {How multilingual is
  multilingual {BERT}?}
\newblock In \emph{Proceedings of the 57th Annual Meeting of the Association
  for Computational Linguistics}, pages 4996--5001, Florence, Italy.
  Association for Computational Linguistics.

\bibitem[{Prasanna et~al.(2020)Prasanna, Rogers, and
  Rumshisky}]{prasanna-etal-2020-bert}
Sai Prasanna, Anna Rogers, and Anna Rumshisky. 2020.
\newblock \href {https://doi.org/10.18653/v1/2020.emnlp-main.259} {{W}hen
  {BERT} {P}lays the {L}ottery, {A}ll {T}ickets {A}re {W}inning}.
\newblock In \emph{Proceedings of the 2020 Conference on Empirical Methods in
  Natural Language Processing (EMNLP)}, pages 3208--3229, Online. Association
  for Computational Linguistics.

\bibitem[{Shapley(1953)}]{shapley_1953}
Lloyd~S. Shapley. 1953.
\newblock \href {https://doi.org/10.1017/CBO9780511528446.003} {\emph{A value
  for n-person games}}, page 31–40. Cambridge University Press.

\bibitem[{Sundararajan et~al.(2017)Sundararajan, Taly, and
  Yan}]{integratedGradients}
Mukund Sundararajan, Ankur Taly, and Qiqi Yan. 2017.
\newblock \href {http://proceedings.mlr.press/v70/sundararajan17a.html}
  {Axiomatic attribution for deep networks}.
\newblock In \emph{Proceedings of the 34th International Conference on Machine
  Learning, {ICML} 2017, Sydney, NSW, Australia, 6-11 August 2017}, volume~70
  of \emph{Proceedings of Machine Learning Research}, pages 3319--3328. {PMLR}.

\bibitem[{Tiedemann et~al.(2014)Tiedemann, Agi{\'c}, and
  Nivre}]{tiedemann-etal-2014-treebank}
J{\"o}rg Tiedemann, {\v{Z}}eljko Agi{\'c}, and Joakim Nivre. 2014.
\newblock \href {https://doi.org/10.3115/v1/W14-1614} {Treebank translation for
  cross-lingual parser induction}.
\newblock In \emph{Proceedings of the Eighteenth Conference on Computational
  Natural Language Learning}, pages 130--140, Ann Arbor, Michigan. Association
  for Computational Linguistics.

\bibitem[{Vig(2019)}]{vig-2019-multiscale}
Jesse Vig. 2019.
\newblock \href {https://doi.org/10.18653/v1/P19-3007} {A multiscale
  visualization of attention in the transformer model}.
\newblock In \emph{Proceedings of the 57th Annual Meeting of the Association
  for Computational Linguistics: System Demonstrations}, pages 37--42,
  Florence, Italy. Association for Computational Linguistics.

\bibitem[{Voita et~al.(2019)Voita, Talbot, Moiseev, Sennrich, and
  Titov}]{voita-etal-2019-analyzing}
Elena Voita, David Talbot, Fedor Moiseev, Rico Sennrich, and Ivan Titov. 2019.
\newblock \href {https://doi.org/10.18653/v1/P19-1580} {Analyzing multi-head
  self-attention: Specialized heads do the heavy lifting, the rest can be
  pruned}.
\newblock In \emph{Proceedings of the 57th Annual Meeting of the Association
  for Computational Linguistics}, pages 5797--5808, Florence, Italy.
  Association for Computational Linguistics.

\bibitem[{Wang et~al.(2020)Wang, Lipton, and
  Tsvetkov}]{wang-etal-2020-negative}
Zirui Wang, Zachary~C. Lipton, and Yulia Tsvetkov. 2020.
\newblock \href {https://doi.org/10.18653/v1/2020.emnlp-main.359} {On negative
  interference in multilingual models: Findings and a meta-learning treatment}.
\newblock In \emph{Proceedings of the 2020 Conference on Empirical Methods in
  Natural Language Processing (EMNLP)}, pages 4438--4450, Online. Association
  for Computational Linguistics.

\bibitem[{Wiegreffe and Pinter(2019)}]{attention-is-not-not}
Sarah Wiegreffe and Yuval Pinter. 2019.
\newblock \href {https://doi.org/10.18653/v1/D19-1002} {Attention is not not
  explanation}.
\newblock In \emph{Proceedings of the 2019 Conference on Empirical Methods in
  Natural Language Processing and the 9th International Joint Conference on
  Natural Language Processing (EMNLP-IJCNLP)}, pages 11--20, Hong Kong, China.
  Association for Computational Linguistics.

\bibitem[{Xia et~al.(2022)Xia, Zhong, and Chen}]{compact}
Mengzhou Xia, Zexuan Zhong, and Danqi Chen. 2022.
\newblock \href {https://aclanthology.org/2022.acl-long.107} {Structured
  pruning learns compact and accurate models}.
\newblock In \emph{Proceedings of the 60th Annual Meeting of the Association
  for Computational Linguistics (Volume 1: Long Papers), {ACL} 2022, Dublin,
  Ireland, May 22-27, 2022}, pages 1513--1528. Association for Computational
  Linguistics.

\bibitem[{Xue et~al.(2021)Xue, Constant, Roberts, Kale, Al-Rfou, Siddhant,
  Barua, and Raffel}]{mt5-xue}
Linting Xue, Noah Constant, Adam Roberts, Mihir Kale, Rami Al-Rfou, Aditya
  Siddhant, Aditya Barua, and Colin Raffel. 2021.
\newblock \href {https://doi.org/10.18653/v1/2021.naacl-main.41} {m{T}5: A
  massively multilingual pre-trained text-to-text transformer}.
\newblock In \emph{Proceedings of the 2021 Conference of the North American
  Chapter of the Association for Computational Linguistics: Human Language
  Technologies}, pages 483--498, Online. Association for Computational
  Linguistics.

\bibitem[{Zeman(2008)}]{zeman2008reusable}
Daniel Zeman. 2008.
\newblock Reusable tagset conversion using tagset drivers.
\newblock In \emph{Proceedings of the Sixth International Conference on
  Language Resources and Evaluation (LREC'08)}.

\bibitem[{Zhao and Sch{\"u}tze(2021)}]{zhao-schutze-2021-discrete}
Mengjie Zhao and Hinrich Sch{\"u}tze. 2021.
\newblock \href {https://doi.org/10.18653/v1/2021.emnlp-main.672} {Discrete and
  soft prompting for multilingual models}.
\newblock In \emph{Proceedings of the 2021 Conference on Empirical Methods in
  Natural Language Processing}, pages 8547--8555, Online and Punta Cana,
  Dominican Republic. Association for Computational Linguistics.

\end{thebibliography}

\end{document}